\documentclass{article}

\usepackage{times}
\usepackage{graphicx} 
\usepackage{subfigure} 

\usepackage{natbib}

\usepackage{algorithm}
\usepackage{algorithmic}

\usepackage{hyperref}

\usepackage[accepted]{icml2015}

\icmltitlerunning{
Deep Convolutional Neural Network Inference with
Floating-point Weights and Fixed-point Activations
}

\begin{document} 

\twocolumn[
\icmltitle{
Deep Convolutional Neural Network Inference with
Floating-point Weights and Fixed-point Activations
}

\icmlauthor{Liangzhen Lai}{liangzhen.lai@arm.com}
\icmladdress{ARM Research,
             150 Rose Orchard Way,
             San Jose, CA 95134 USA}
\icmlauthor{Naveen Suda}{naveen.suda@arm.com}
\icmladdress{ARM Research,
             150 Rose Orchard Way,
             San Jose, CA 95134 USA}
\icmlauthor{Vikas Chandra}{vikas.chandra@arm.com}
\icmladdress{ARM Research,
             150 Rose Orchard Way,
             San Jose, CA 95134 USA}

\icmlkeywords{machine learning}

\vskip 0.3in
]

\begin{abstract} 
Deep convolutional neural network (CNN) inference requires significant amount
of memory and computation, which limits its deployment on embedded devices. 
To alleviate these problems to some extent, prior research utilize low precision 
fixed-point numbers to represent the CNN weights 
and activations. However, the minimum required data precision of fixed-point weights 
varies across different networks and also across different layers of the same 
network. In this work, we propose using floating-point numbers for representing
the weights and fixed-point numbers for representing the activations. We 
show that using floating-point representation for weights is more efficient than 
fixed-point representation for the same bit-width and demonstrate it on popular 
large-scale CNNs such as AlexNet, SqueezeNet, GoogLeNet and VGG-16.
We also show that such a representation scheme enables compact hardware 
multiply-and-accumulate (MAC) unit design. Experimental results show that the 
proposed scheme reduces the weight storage by up to 36\% and power consumption
of the hardware multiplier by up to 50\%.
\end{abstract} 

\section{Introduction}
\label{sec:introduction}

Deep learning spearheaded by convolutional neural networks (CNN) and recurrent neural 
networks (RNN) has been pushing the frontiers in many computer vision 
applications~\cite{lecun2015deep}. Deep learning algorithms have achieved or surpassed 
human-levels of perception in some applications~\cite{he2016deep,xiong2016microsoft}, 
enabling them to be deployed in the real-world applications.
The key challenges of deep learning algorithms are the computational complexity and
model size, which impede their deployment on end-user client devices, 
thus limiting them to cloud-based high performance servers. 
For example, AlexNet~\cite{krizhevsky2012imagenet}, a popular CNN model, requires 1.45 billion 
operations per input image with 240MB weights~\cite{suda2016throughput}.

Many researchers have explored hardware accelerators for CNNs to enable deployment 
of CNNs in embedded devices and demonstrated good performance at low power 
consumption~\cite{qiu2016going,shin2017dnpu}. Reducing the data precision is a
commonly used technique for improving the energy efficiency of CNNs.
Typically, CNNs are trained on high performance CPU/GPU with 32-bit floating-point data.
Fixed-point representation with shorter bit-width for CNN weights and 
activations has been widely
explored~\cite{judd2015reduced,gupta2015deep,gysel2016hardware,lin2015fixed},
which significantly reduces the storage requirements, memory 
bandwidth and power consumption without sacrificing accuracy.

This work focuses on the number representation schemes for implementing
CNN inference. The representation scheme has the following requirements:
\begin{itemize}
\item Accuracy: the representation should achieve the desired network accuracy
with limited bit-width.
\item Efficiency: the representation can be implemented in hardware efficiently.
\item Consistency: the representation should be consistent across different CNNs.
\end{itemize}

Based on these requirements, we propose using floating-point numbers for CNN weights 
and fixed-point numbers for activations. We justify this choice from both 
algorithmic and hardware implementation perspectives. 
From the algorithmic perspective, using popular large-scale CNNs 
such as AlexNet~\cite{krizhevsky2012imagenet}, 
SqueezeNet~\cite{iandola2016squeezenet}, GoogLeNet~\cite{szegedy2015going}
and VGG-16~\cite{simonyan2014very}, 
we show that the representation range of the weights is the main factor that 
determines the inference accuracy, which can be better represented
in floating-point format.
From the hardware perspective, we show that multiplication 
can be implemented more efficiently with one floating-point operand and one 
fixed-point operand generating a fixed-point product.


The rest of the paper is organized as follows, Section~\ref{sec:related_work}
discusses related work. Section~\ref{sec:background} gives some background
about different number representation formats and typical hardware implementation
of CNNs. Section~\ref{sec:scheme} describes the proposed number representation
scheme. Section~\ref{sec:algorithmic} and Section~\ref{sec:implementation} 
discuss the scheme from the algorithmic and implementation perspective.
Section~\ref{sec:results} presents the experimental results, and 
Section~\ref{sec:conclusion} concludes the paper.

\section{Related Work}
\label{sec:related_work}

Precision of the neural network weights and activations plays a major role in 
determining the efficiency of the CNN hardware or software implementations. 
A lot of research focuses on replacing the standard 32-bit floating-point 
data with reduced precision data for CNN inference.
For example, Gysel et al.~\cite{gysel2016ristretto} propose representing
both CNN weights and activations using minifloat, i.e., floating-point number
with shorter bit-width.
Since fixed-point arithmetic is more hardware efficient than floating-point 
arithmetic, most research focuses on fixed-point quantization.
Gupta et al.~\cite{gupta2015deep} present the impacts of different fixed-point 
rounding schemes on the 
accuracy. Judd et al.~\cite{judd2015reduced} demonstrate that the minimum 
required data precision not only 
varies across different networks, but also across different layers of the 
same network. Lin et al.~\cite{lin2015fixed} 
present a fixed-point quantization methodology to identify the optimal data 
precision for all layers of a network. 
Gysel et al.~\cite{gysel2016hardware} present a framework
{\em{Ristretto}} for fixed-point quantization and re-training of CNNs based on
{\em{Caffe}}~\cite{jia2014caffe}. 

Researchers have also explored training neural networks directly 
with fixed-point weights. In~\cite{hammerstrom1990vlsi}, the author 
presents a hardware architecture for on-chip learning with 
fixed-point operations. More recently, in~\cite{courbariaux2014training}, 
the authors train neural networks with floating-point, fixed-point and 
dynamic fixed-point formats and demonstrate that fixed-point 
weights are sufficient for training. Gupta et al.~\cite{gupta2015deep} demonstrate network 
training with 16-bit fixed-point weights using stochastic rounding scheme.

Many other approaches for memory reduction of neural networks have been explored. 
Han et al.~\cite{han2015deep} 
propose a combination of network pruning, weight quantization during training 
and compression based on Huffman 
coding to reduce the VGG-16 network size by 49X. 
In~\cite{deng2015reduced}, the authors propose to store both 8-bit quantized 
floating-point weights and 32-bit full precision weights. At runtime, quantized weights 
or full-precision weights are randomly fetched in order to reduce memory bandwidth. 
The continuous research effort to reduce the data precision 
has led to many interesting demonstrations with 2-bit weights~\cite{intel2bitnn} 
and even binary weights/activations~\cite{binarynet,xnornet}. 
Zhou et al.~\cite{dorefanet} demonstrate AlexNet
training with 1-bit weights, 2-bit activations and 6-bit gradients.
These techniques require additional re-training and can result in sub-optimal
accuracies.

In contrast to prior works, this work proposes quantization of a pre-trained 
neural network weights into floating-point numbers and implementation of
activations in fixed-point format both for 
memory reduction and hardware 
efficiency. It further shows that floating-point representation of weights 
achieves better range/accuracy trade-off 
compared for the fixed-point representation of same number of bits and 
we empirically demonstrate it on state of the art
CNNs such as AlexNet~\cite{krizhevsky2012imagenet}, VGG-16~\cite{simonyan2014very}, 
GoogLeNet~\cite{szegedy2015going} 
and SqueezeNet~\cite{iandola2016squeezenet}. 
Although this work is based on quantization only without 
the need for retraining the network, retraining may also be applied to
reclaim part of the accuracy loss due to quantization.

\section{Background}
\label{sec:background}

\subsection{Fixed-Point Number Representation}
\label{subsec:background_fixed_point}
\begin{figure}[t]
\centering
\includegraphics[width = 0.95\columnwidth]{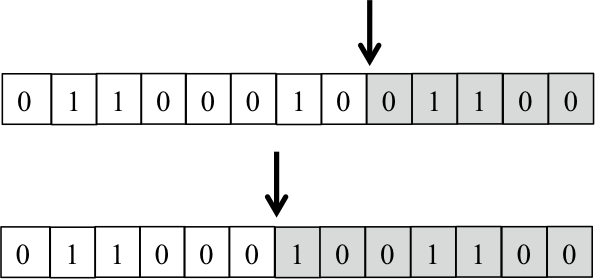}
\caption{
Examples of fixed-point representation. The arrow indicates the position of
radix point. So the first number is the corresponding integer value scaled by
$2^{-5}$ and the second one is scaled by $2^{-7}$.
}
\label{fig:fixed-point}
\end{figure}

Fixed-point representation is very similar to integer representation. The 
difference is that integer has a scaling factor of 1 and fixed-point can have a 
pre-defined scaling factor as power of 2. 
Some examples of fixed-point numbers are shown in Fig.~\ref{fig:fixed-point}.

Usually, all fixed-point representation is assumed to share the same scaling 
factor during the entire computation. In some scenarios, the computation can 
be classified into different sets, e.g., for different CNN layers, with 
fixed-point numbers of different scaling factors. This is also referred as 
dynamic fixed-point 
representation~\cite{courbariaux2014training}.

\subsection{Floating-point Number Representation}
\label{subsec:background_floating_point}

\begin{figure}[t]
\centering
\includegraphics[width = 0.95\columnwidth]{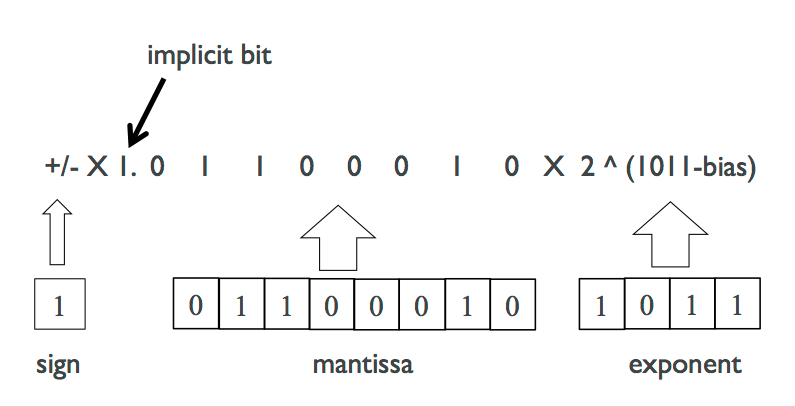}
\caption{
Example of floating-point representation.
With the implicit bit, the range for the significand part is limited to be $[1, 2)$.
Without the implicit bit, i.e., leading bit as 0, the range becomes
$[0, 0.5)$.
}
\label{fig:floating-point}
\end{figure}

One example of floating-point number representation is shown in
Fig.~\ref{fig:floating-point}. 
For a floating-point representation, there are typically three parts: sign, 
mantissa and exponent. The sign bit determines whether the number is a positive 
or negative number. The mantissa determines the significand part and the exponent 
determine the scale of the value. Usually, there are some special encodings used 
for representing some special numbers (e.g., 0, NaN and +/- infinity), 

For binary floating-point numbers, the mantissa can assume an implicit bit, which 
is also adopted by IEEE floating-point standard. This ensures that the value of 
mantissa is always between 1 and 2, so the leading bit 1 can be omitted to save 
storage space. However, such an implicit bit places a limit on the smallest
representable number for the significand part.`

The exponent is typically represented as an unsigned integer number with a bias. 
For example, for an 8-bit exponent with a bias of 127, it can represent numbers from 
-127 to 128, i.e, 0-127 to 255-127.

\subsection{Hardware Implementation of CNNs}
\label{subsec:background_cnn_hw}

CNNs typically consist of multiple convolution layers interspersed by pooling, ReLU and
normalization layers followed by fully-connected layers. Convolution and 
fully-connected layers are the most compute and data intensive layers 
respectively~\cite{qiu2016going}. 
The computation in these layers consist of multiply-and-accumulate (MAC) operations.
The data path is illustrated in Fig.~\ref{fig:data_flow}, 
where the input features are multiplied with the weights to get the intermediate data 
(i.e., partial sums). These partial sums are accumulated to generate the output features.
Since fixed-point arithmetic is typically more efficient for hardware implementation,
most hardware accelerators implement the MAC operations using fixed-point representation.

The power/area breakdown of the CNN hardware accelerator mostly depends on the data flow 
architecture. For example, in Eyeriss~\cite{chen2016eyeriss}, in each processing element (PE),
MAC and memory account for about 9\% and 52\% area respectively. 
For Origami~\cite{cavigelli2015origami}, MAC and memory account for about
32\% and 34\% area respectively.

\section{Proposed Number Representation Scheme}
\label{sec:scheme}

\begin{figure}[t]
\centering
\includegraphics[width = 0.95\columnwidth]{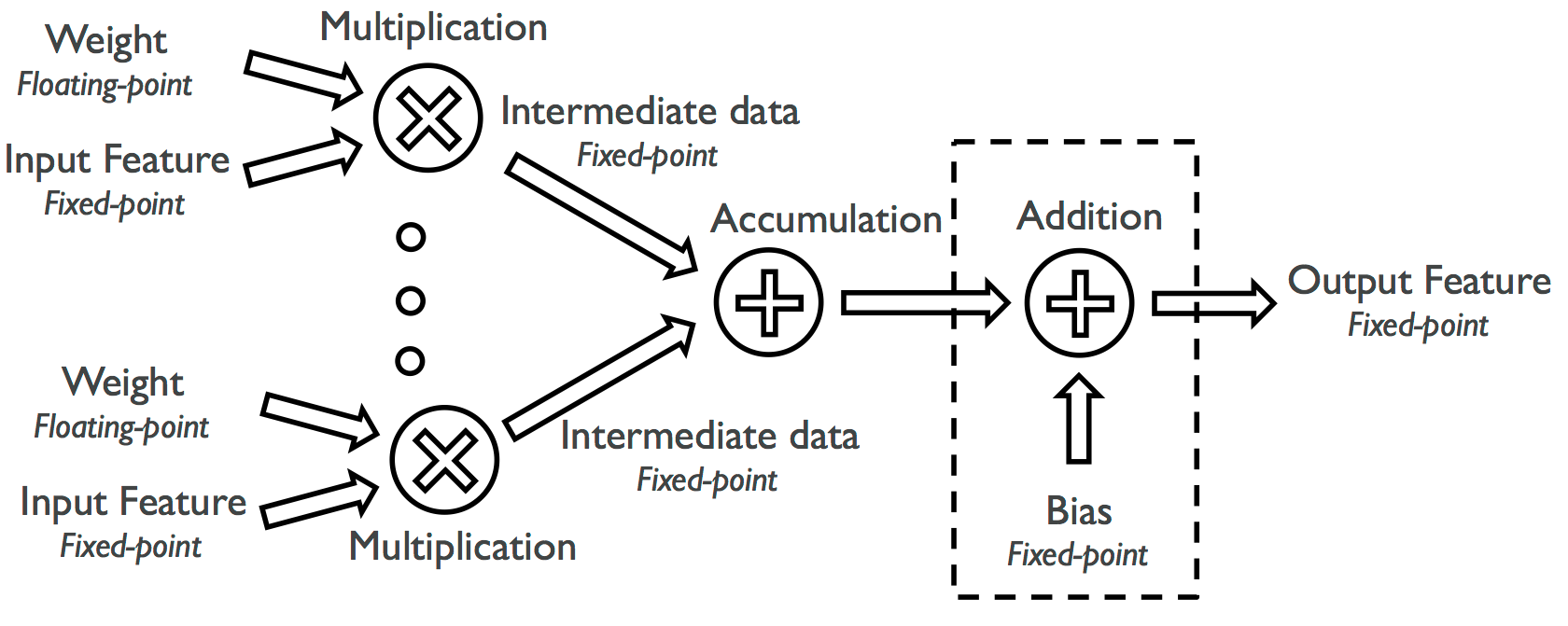}
\caption{
Overview of the data path for convolutional and fully-connected layer
operations with the proposed number representation scheme.
The weights are represented using floating-point numbers, while the
input/output features and intermediate data are represented using
fixed-point numbers.
}
\label{fig:data_flow}
\end{figure}

The overview of our proposed number representation scheme is shown in
Fig.~\ref{fig:data_flow}. Different from most existing CNN implementations,
we propose using a combination of floating-point and fixed-point
representations.
The network weights are represented as floating-point numbers while the 
input/output features are represented as fixed-point numbers.
The multiplier is implemented to take one floating-point number and one
fixed-point number and produces output, i.e., intermediate data, in 
fixed-point format.
The intermediate data is in fixed-point format and can have wider bit-width 
than the input/output features. 
The accumulation is the same as fixed-point adder, which can have higher
bit-width. 

From hardware perspective, it is more efficient to implement multiplication using 
floating-point number and addition using fixed-point number. 
The multiplication operations have one fixed-point number input, one floating-point 
number input and fixed-point number output. 
This can be implemented with a multiplier and a shifter. 
The multiplier will multiply the fixed-point number with the mantissa part of the 
floating-point number, and the shifter will shift the results according to the 
exponent value. Therefore, we propose the hardware architecture illustrated in 
Fig.~\ref{fig:data_flow}.

The accumulation/addition works with fixed-point numbers, which can be wider 
(i.e., with larger bit-width) than either of the inputs. This part is similar to 
most fixed-point number based implementation of CNN accelerators. 

\section{Algorithmic Perspective}
\label{sec:algorithmic}

In this section, we investigate and explain why the proposed number representation 
scheme is better from the algorithmic perspective.
Section~\ref{subsec:fixed_point_accuracies} demonstrates that different CNNs can have
inconsistent fixed-point bit-width requirements for representing the weights.
Section~\ref{subsec:weight_distribution} investigates this inconsistency by
analyzing CNN weight distribution and properties.
Section~\ref{subsec:range_vs_precision} shows that the representation range is the 
main factor that determines the inference accuracy.
Section~\ref{subsec:floating_point_accuracies} shows that floating-point representation
is more efficient and consistent representation for CNN weights.

\subsection{CNN Accuracy with Fixed-Point Weights}
\label{subsec:fixed_point_accuracies}

To evaluate different number representation schemes, we implement
weight quantization based on {\em{Caffe}}~\cite{jia2014caffe} framework.
To make a fair comparison, we assume that there is always a sign bit for representing
negative values for both fixed-point and floating-point numbers.
We will represent the bit-width of floating point representation as $m+e$, where
$m$ is the number of mantissa bits and $e$ is the number of exponent bits.

We apply the weight quantization on four popular CNN networks:
AlexNet~\cite{krizhevsky2012imagenet}, 
SqueezeNet~\cite{iandola2016squeezenet}, GoogLeNet~\cite{szegedy2015going}
and VGG-16~\cite{simonyan2014very}. 
We evaluate the network accuracy by doing quantization for all convolutional and
fully-connected layer weights. The activation is quantized to 16-bit fixed-point.
For each layer, we normalize the weights so that the maximum absolute value equals 1.
This is similar to the dynamic fixed-point
quantization~\cite{moons2016energy,gysel2016hardware,courbariaux2014training}.
All accuracy results are top-1 accuracy and normalized with the top-1 accuracy 
using 32-bit floating-point representation. 
The results of top-5 accuracy correlate with that of top-1 accuracy.

\begin{figure}[t]
\centering
\includegraphics[width = 0.95\columnwidth]{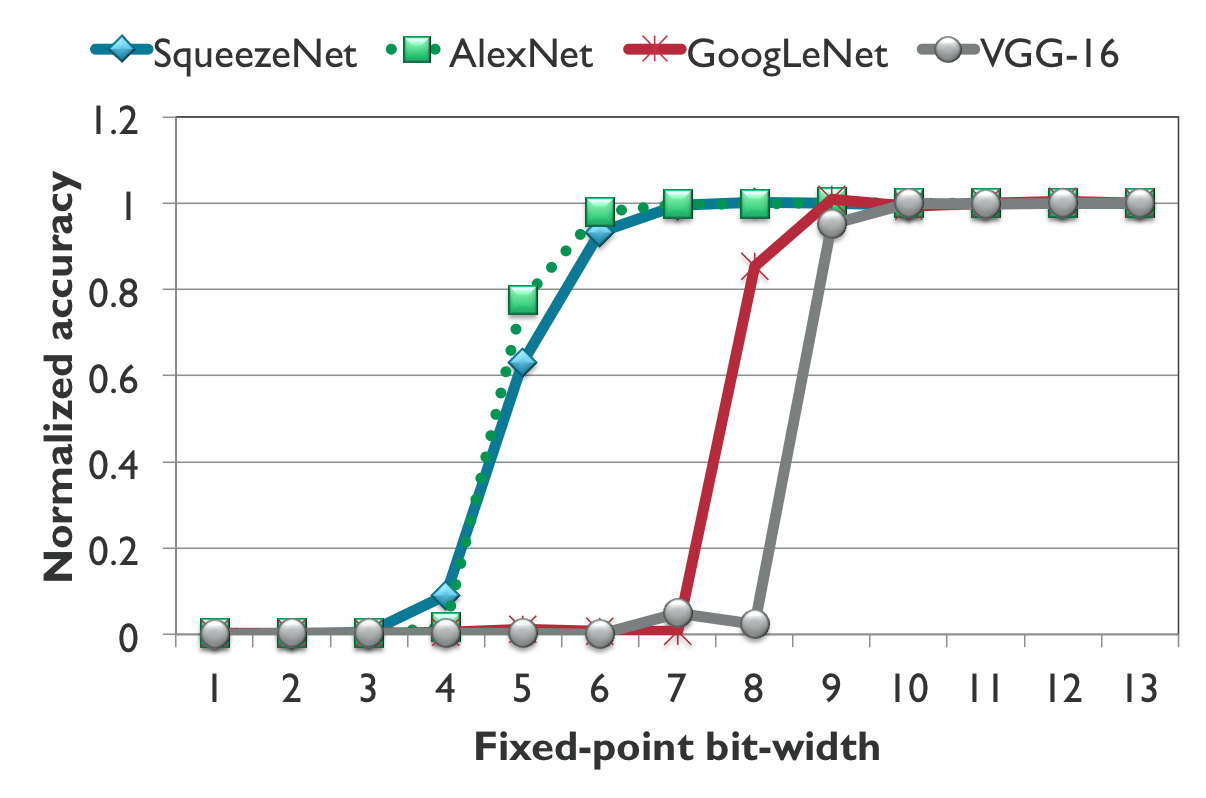}
\caption{Normalized accuracy of the networks using fixed-point representation with
different bit-width.}
\label{fig:acc_fixed}
\end{figure}

The network accuracy results using fixed-point representation are shown in 
Fig.~\ref{fig:acc_fixed}. Two observations can be made here:
\begin{itemize}
\item For all networks, the accuracy starts increasing sharply after a certain 
threshold. For all networks, the normalized accuracy increases
from close to 0 to close to 1 within 2 or 3 bits difference. This suggests that
there is something dramatically different with these additional bits.

\item Among different networks, the required number of bits is very inconsistent.
With 7-bit fixed-point number, AlexNet and SqueezeNet can achieve close to
full accuracy, while GoogLeNet and VGG-16 have very low accuracy.
GoogLeNet and VGG-16 need 10 to 11 bits to achieve full accuracy.
\end{itemize}

This inconsistency in bit-width requirements across different CNNs
poses challenges for hardware implementation.
For the design to be general-purpose and future-proof, the designer has to use
margined bit-width or use runtime adaptation~\cite{moons2016energy},
both of which incur significant overhead.

\subsection{Weight Distribution}
\label{subsec:weight_distribution}

There can be several reasons that cause the inconsistency in Fig.~\ref{fig:acc_fixed}.
The network depth is one of the possible reasons. Similar to the idea
in~\cite{lin2015fixed}, the fixed-point quantization can be modeled as quantization
noise for each layer. The network accuracy may drop further, i.e., accumulate
more noise, as the network depth increases.
The other possible reason is the number of MAC operations in each layer. 
Small quantization error can accumulate over a large amount of MAC operations.
For example, the total number of MAC operations to calculate one output for convolutional and
fully-connected layers in AlexNet are (363, 1200, 2304, 1728, 1728, 9216,
4096, 4096).

\begin{figure}[t]
\centering
\includegraphics[width = 0.95\columnwidth]{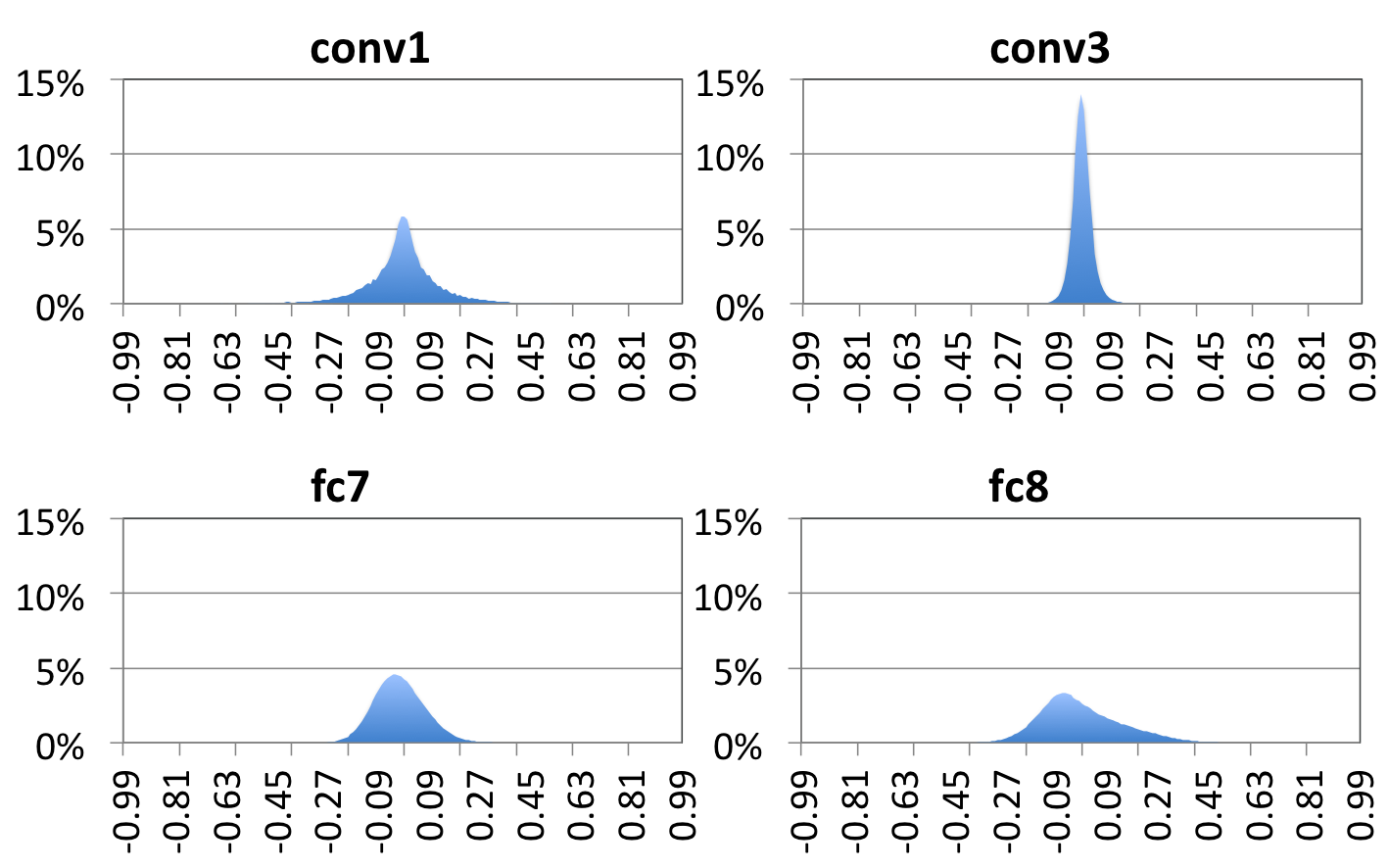}
\caption{
The weight distribution of four different layers in AlexNet.
{\em{conv1}} and {\em{conv3}} are convolutional layers, and fc7 and fc8 are fully-connected
layers.
}
\label{fig:alex_lin_hist}
\end{figure}

However, none of the above reasons can explain the first observation earlier about
the sharp, instead of gradual, change in accuracy.
To further investigate this, we plot the weight distribution of four different layers
in AlexNet in Fig.~\ref{fig:alex_lin_hist}.
Most weights have small values even after the per-layer normalization. The distribution
is concentrated at the center, which is also the motivation for Huffman encoding 
of the weights proposed in~\cite{han2015deep}.

\begin{figure}[t]
\centering
\includegraphics[width = 0.95\columnwidth]{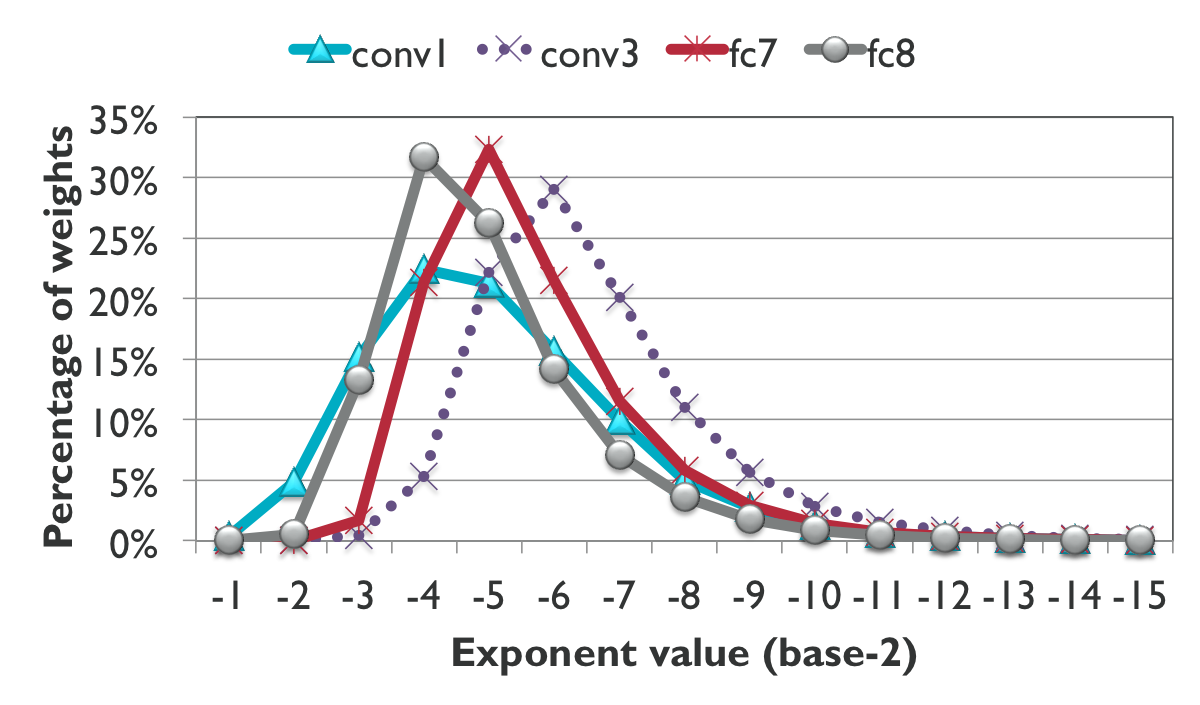}
\caption{
The weight distribution in log-scale of four different layers in AlexNet.
The weights are first converted into binary floating-point format. The plot
is based on the exponent value of the weights.
}
\label{fig:alex_hist}
\end{figure}

To better visualize the difference, we plot the same weight distribution in log-scale
in Fig.~\ref{fig:alex_hist}. 
This plot is easier to spot the weight distribution difference and explains the accuracy
behavior under fixed-point quantization observed in Fig.~\ref{fig:acc_fixed}.
Under fixed-point quantization, the layer with the most small-valued weights, {\em{conv3}},
will be the most susceptible to quantization errors.
With 4-bit fixed-point representation, more than 90\%
weights in {\em{conv3}} are unrepresentable, i.e., quantized to 0.
This also explains why stochastic rounding works better than round-to-nearest reported 
in~\cite{gupta2015deep}.

Since the layers are cascaded, the accuracy of the entire network is limited
by the weakest layer. This is why the network produces close to 0 accuracy with small
bit-width as shown in Fig.~\ref{fig:acc_fixed}.
With higher fixed-point bit-width, a larger number of weights in {\em{conv3}} become 
representable, which results in the quick increase of the network accuracy.

\begin{figure}[t]
\centering
\includegraphics[width = 0.95\columnwidth]{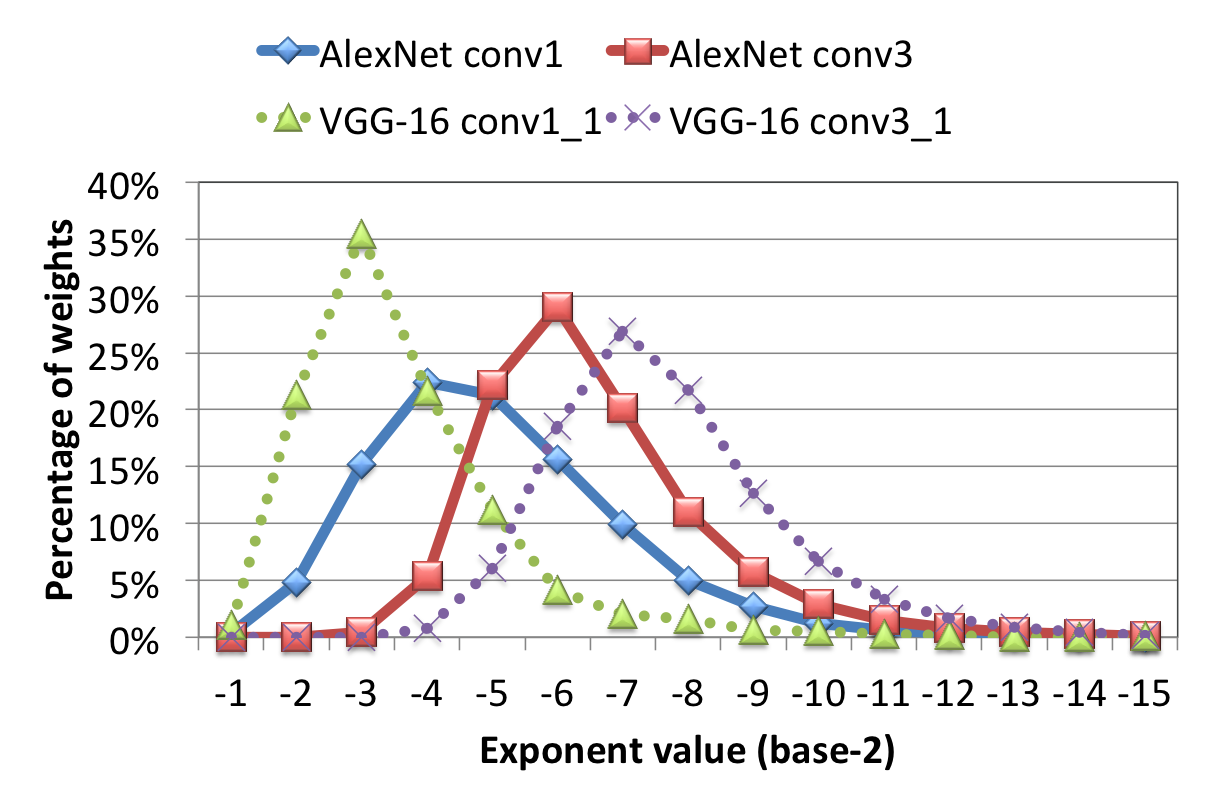}
\caption{
The weight distribution in log-scale for different layers in AlexNet and VGG-16.
For each network, we pick the layers with the largest and smallest mean in
the exponent value of the weights.
}
\label{fig:hist_comparison}
\end{figure}

The inconsistency in bit-width requirements observed in 
Fig.~\ref{fig:acc_fixed} can also be explained with
the weight distribution. We pick the two layers with largest and smallest
weights from AlexNet and VGG-16 and plot the weight distribution in
Fig.~\ref{fig:hist_comparison}.
The layer {\em{conv3$\_$1}} in VGG-16 has more weights with smaller exponent values.
This explains why VGG-16 requires more bit-width when using fixed-point representation.

\subsection{Range vs. Precision}
\label{subsec:range_vs_precision}

The results in Fig.~\ref{fig:acc_fixed} and the weight distributions in
Fig.~\ref{fig:alex_hist} show that the network can achieve almost         
full accuracy, e.g., with 7-bit fixed-point for AlexNet, even when most 
weights are barely representable, i.e., only with 1 or 2 significant bits.
This means that representation range, i.e, the ability to represent
larger/smaller values, is more important than representation precision,
i.e., differentiation between nearby values.

\begin{figure}[t]
\centering
\includegraphics[width = 0.95\columnwidth]{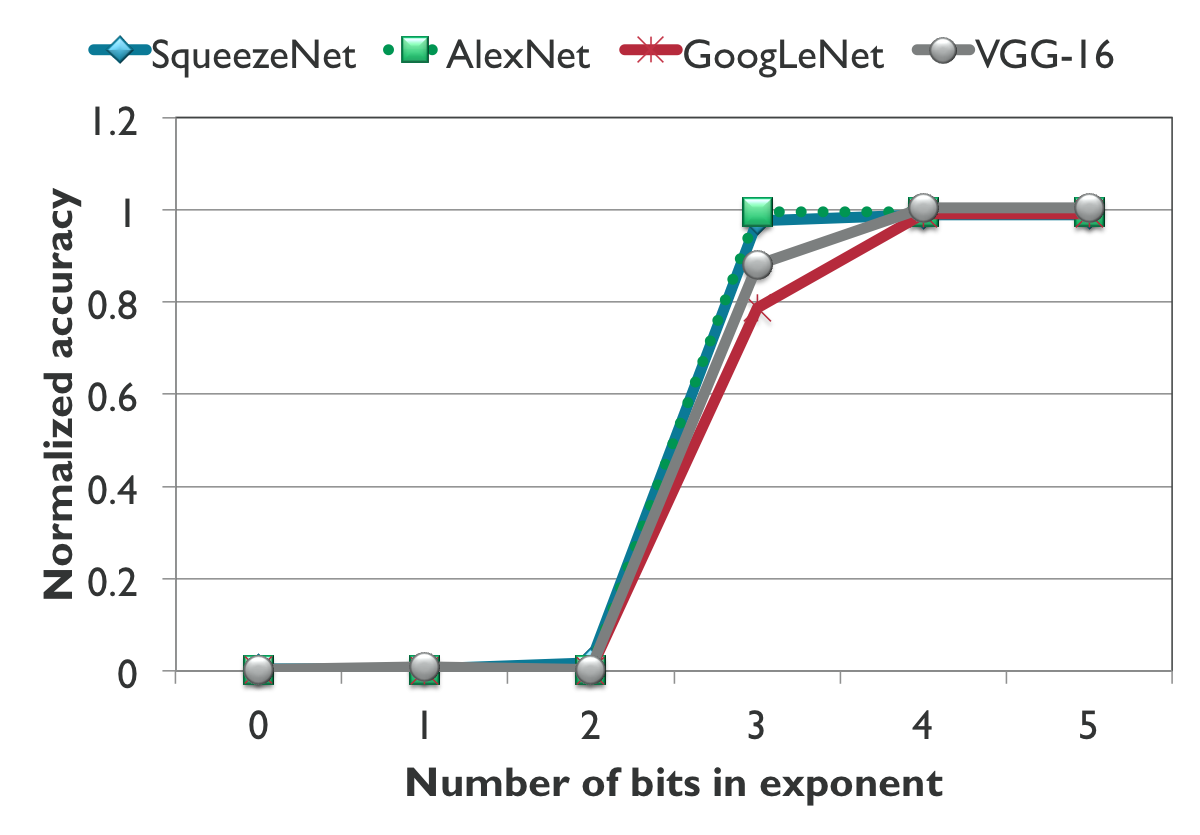}
\caption{Normalized accuracy of the networks using floating-point representation
with different exponent bit-width. The floating-point number has 3-bit mantissa with
the implicit bit.}
\label{fig:acc_float}
\end{figure}

Since the representation range and representation precision is hard to decompose
in fixed-point representation, we investigate this using floating-point
representation.
For floating-point representation, the mantissa bit-width controls the precision
and the exponent bit-width controls the range.

Fig.~\ref{fig:acc_float} highlights some of the results of network accuracy
using floating-point representation with varying exponent bit-width (i.e., 
representation range).
The floating-point has 3-bit mantissa with the implicit bit. The implicit
bit limits the value of the significant part, so that the representation
range is controlled by the exponent part.
With floating-point representation, the networks show consistent trend,
with almost 0 accuracy with 2-bit exponent and quickly increase to almost 
full accuracy with 4-bit exponent.
Unlike the behavior seen in Fig.~\ref{fig:acc_fixed}, 
this is expected as 2 additional bits 
in exponent offers 4X increase in the representation range.

\begin{figure}[t]
\centering
\includegraphics[width = 0.95\columnwidth]{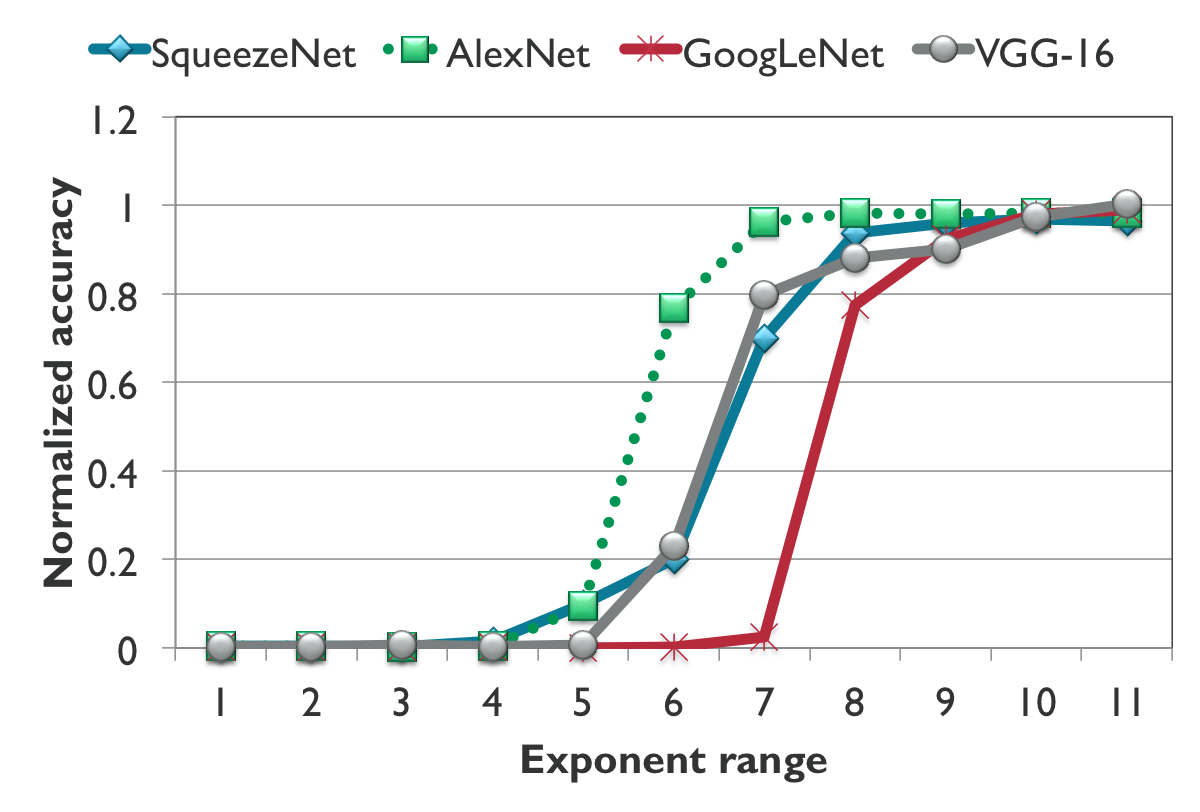}
\caption{Normalized accuracy of the networks using floating-point representation
with limited exponent range. 
Range of 4 is equivalent to 2-bit exponent, and
range of 8 is equivalent to 3-bit exponent.
The floating-point number has 2-bit mantissa with
the implicit bit.}
\label{fig:acc_float_exp2}
\end{figure}

To get more insight into the impact of representation range, we also run 
experiments with floating-point like representation where we limit the 
exponent range rather than the exponent bits. 
For example, exponent range of 4 is equivalent to
2-bit exponent and exponent range of 8 is equivalent to 3-bit exponent.
The results of exponent range experiments are shown in Fig.~\ref{fig:acc_float_exp2}.
The floating-point number has 2-bit mantissa with the implicit bit.
The behavior of the accuracy is similar to the results with fixed-point
representation, i.e., accuracy increases rapidly from almost 0 to almost 1.
The relative ordering exponent range requirements also matches the bit-width
requirements for fixed-point representation. This suggests that representation range is
the main impacting factor for network accuracy.

\begin{figure}[t]
\centering
\includegraphics[width = 0.95\columnwidth]{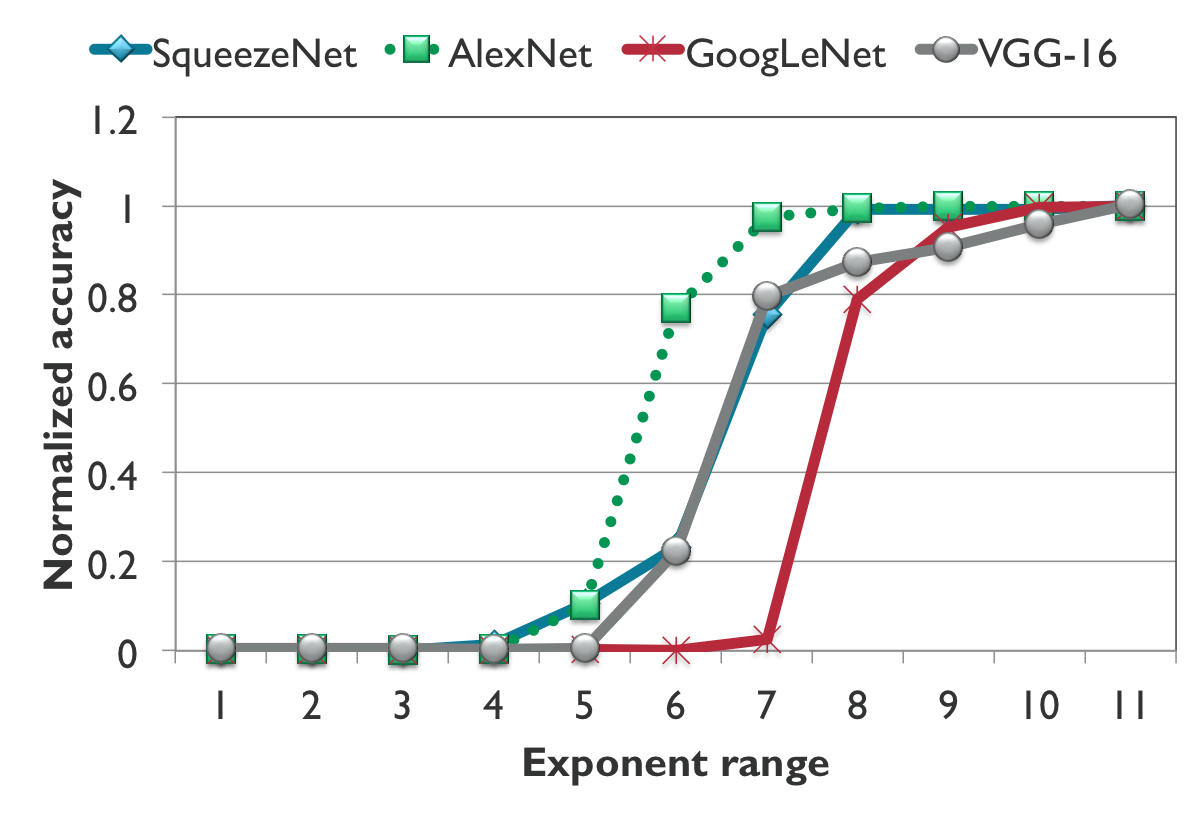}
\caption{Normalized accuracy of the networks using floating-point representation
with limited exponent range.
Range of 4 is equivalent to 2-bit exponent, and
range of 8 is equivalent to 3-bit exponent.
The floating-point number has 6-bit mantissa with
the implicit bit.}
\label{fig:acc_float_exp6}
\end{figure}

With 2-bit mantissa, some networks saturate with normalized accuracy less 
than 1, as shown in Fig.~\ref{fig:acc_float_exp2}.
To compare with the effect of precision, the experiments are repeated
with 6-bit mantissa as shown in Fig.~\ref{fig:acc_float_exp6}.
Comparing Fig.~\ref{fig:acc_float_exp6} with Fig.~\ref{fig:acc_float_exp2},
the additional 4 bits in mantissa does not have significant impact on 
the network accuracy. This also validates the initial hypothesis that
representation range is more important than representation accuracy.

\subsection{CNN Accuracy with Floating-Point Weights}
\label{subsec:floating_point_accuracies}

Comparing to fixed-point representation, floating-point is better for
representation range, which increases exponentially with the exponent bit-width.
The results shown in Fig.~\ref{fig:acc_float} show that 4-bit exponent is adequate
and consistent across different networks.

\begin{figure}[t]
\centering
\includegraphics[width = 0.95\columnwidth]{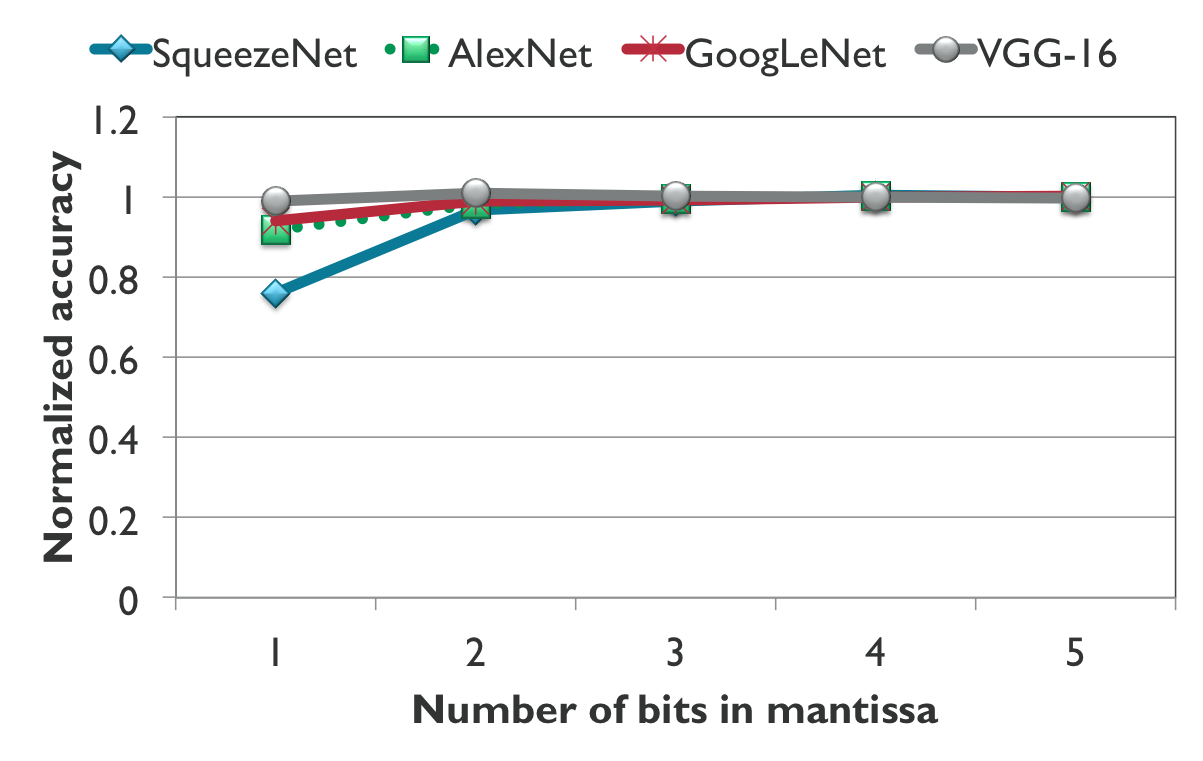}
\caption{Normalized accuracy of the networks using floating-point representation
with different mantissa bit-width. The floating-point number has 4-bit exponent
and is with the implicit bit.}
\label{fig:acc_float_mantissa}
\end{figure}

The next question for floating-point representation is how much precision,
i.e., how many mantissa bits are needed.
Fig.~\ref{fig:acc_float_mantissa} highlights some of the results of network accuracy
using floating-point representation with varying mantissa bit-width.
The floating-point number has 4-bit exponent, and is with the implicit bit.
Most networks have very high accuracy even with 1-bit mantissa and achieve full
accuracy with 3-bit mantissa. This is also consistent across different networks,
which further proves that the inconsistency with fixed-point representation seen 
in Fig.~\ref{fig:acc_fixed}
is mainly from the inconsistent requirements for representation range rather 
than from the representation precision.

\begin{figure}[t]
\centering
\includegraphics[width = 0.95\columnwidth]{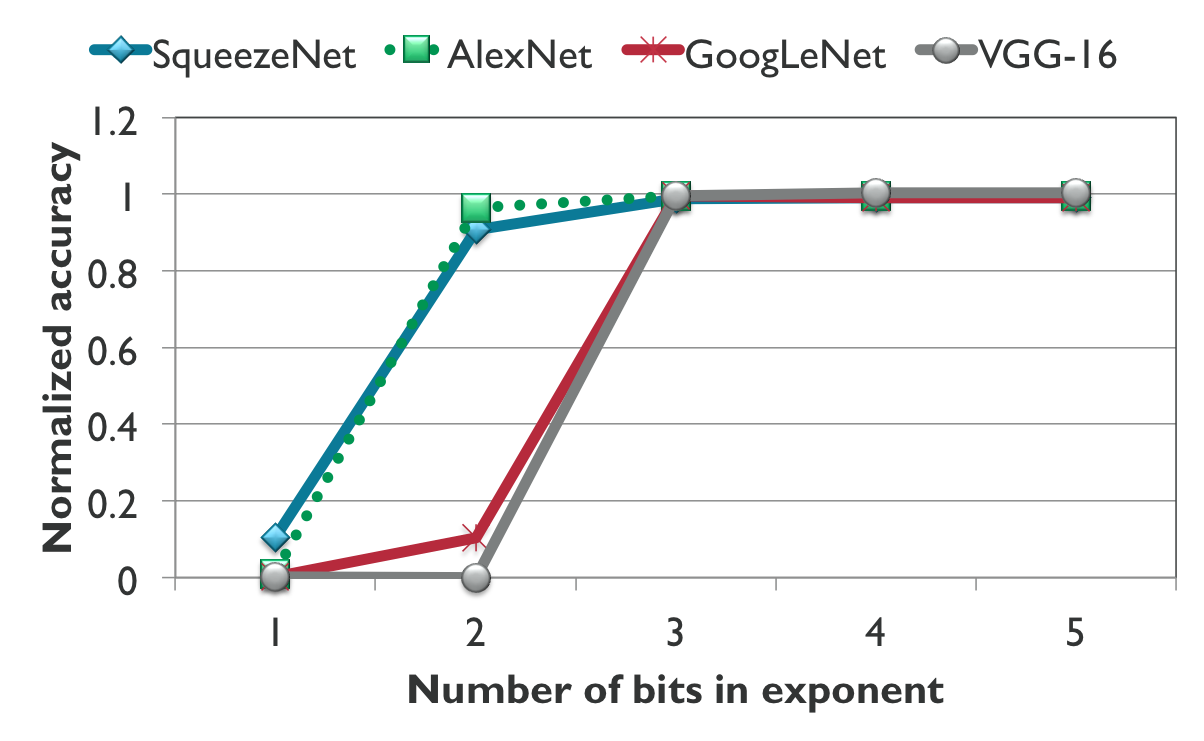}
\caption{Normalized accuracy of the networks using floating-point representation
with different exponent bit-width. The floating-point number has 4-bit mantissa without
the implicit bit.}
\label{fig:acc_float_no}
\end{figure}

\begin{figure}[t]
\centering
\includegraphics[width = 0.95\columnwidth]{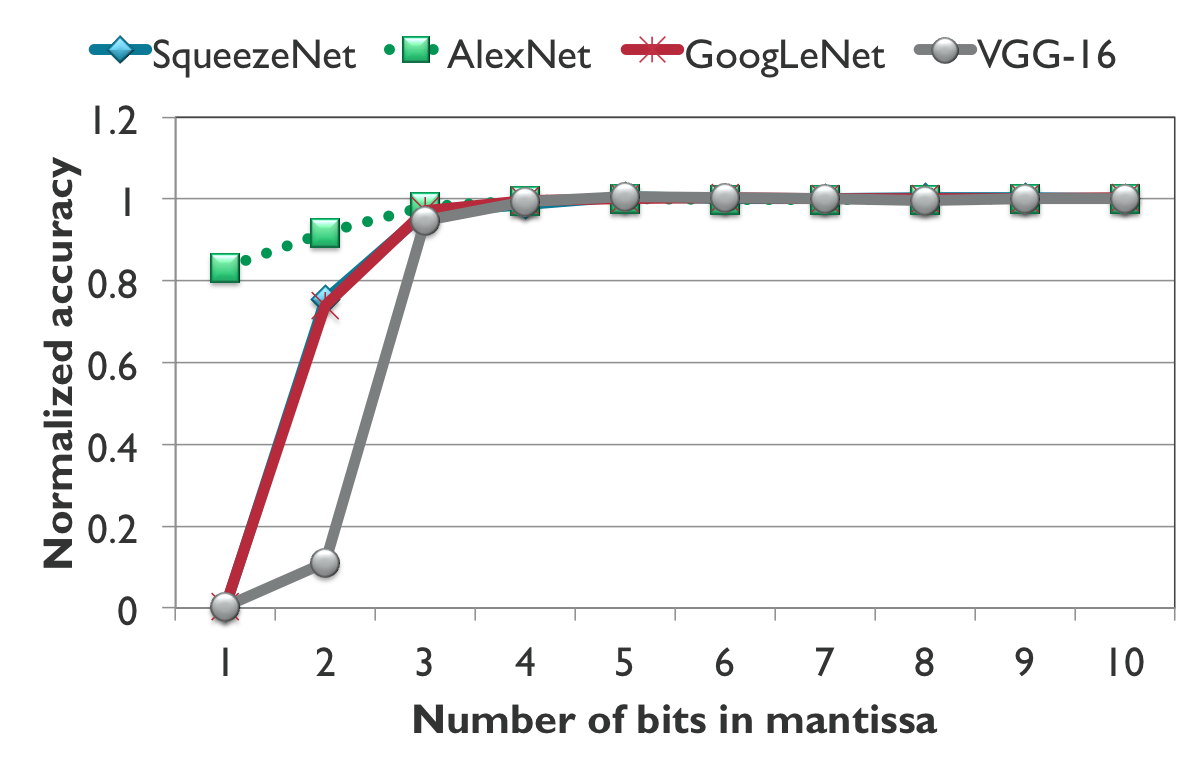}
\caption{Normalized accuracy of the networks using floating-point representation
with different mantissa bit-width. The floating-point number has 3-bit exponent.
The mantissa is without the implicit bit.}
\label{fig:acc_float_mantissa_no}
\end{figure}

We also repeat the experiments with floating-point representation without the
implicit bit in mantissa. The results are highlighted in Fig.~\ref{fig:acc_float_no}
and Fig.~\ref{fig:acc_float_mantissa_no}.
The implicit bit in mantissa limits the range of the significand part to 
$[0.5, 2)$. Removing it helps further extend the range of the representation,
especially for representing small numbers. That is why the network 
accuracy saturates with 3-bit exponent instead of 4-bit.
The implicit bit also improves the precision of the significand part. Hence,
we need 4-bit mantissa instead of 3-bit to achieve full accuracy.

\section{Implementation Perspective}
\label{sec:implementation}

This section motivates the proposed number representation scheme from
the hardware implementation perspective. The implementation considerations 
are discussed in Section~\ref{subsec:implementation_consideration}.
Hardware trade-off results are presented in Section~\ref{subsec:hw_results}.

\subsection{Hardware Implementation Considerations}
\label{subsec:implementation_consideration}

As discussed in Section~\ref{subsec:background_cnn_hw}, computations in CNNs
are typically implemented as MAC operations. 
For the same 32-bit wide operations, hardware implementation of fixed-point 
arithmetic can be more efficient than floating-point arithmetic. 
This is one of the reasons why most of previous work focuses on the optimization
for fixed-point representation.

\begin{figure}[t]
\centering
\includegraphics[width = 0.95\columnwidth]{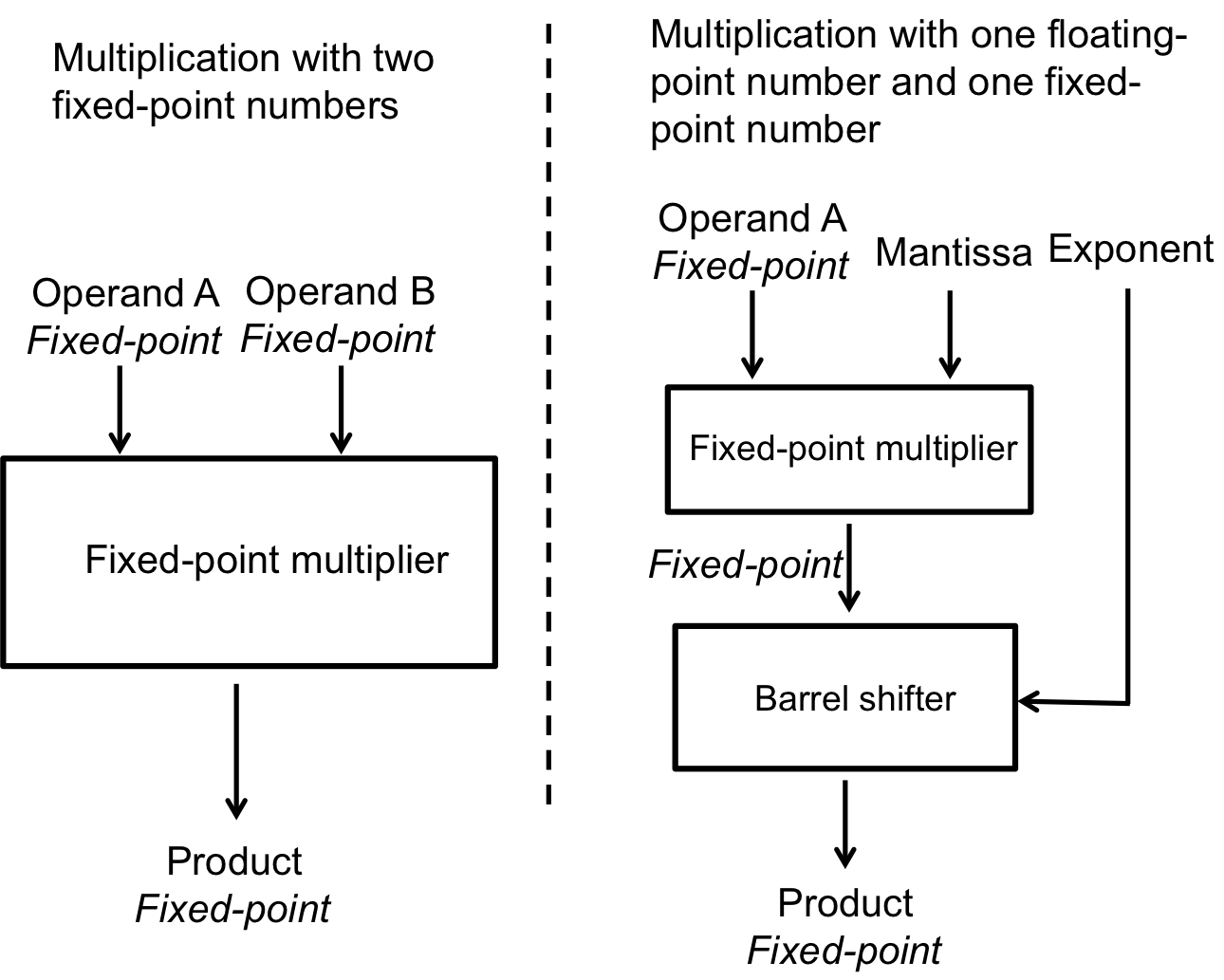}
\caption{
Illustration of hardware implementation of multiplication with two
fixed-point numbers and mixed fixed-point and floating-point numbers.
}
\label{fig:multiplier}
\end{figure}

The comparison becomes less obvious when the bit-width is smaller, especially
when the number of exponent bits in floating-point representation is small.
For example, as shown in Fig.~\ref{fig:multiplier},
multiplier with a floating-point number can be implemented with
a multiplier (of the bit-width of mantissa) and a barrel shifter (of the bit-width of
exponent). This can be more efficient than multiplying two fixed-point numbers,
as the multiplier becomes smaller and the shifter is simpler.

\subsection{Hardware Trade-off Results}
\label{subsec:hw_results}

\begin{figure}[t]
\centering
\includegraphics[width = 0.95\columnwidth]{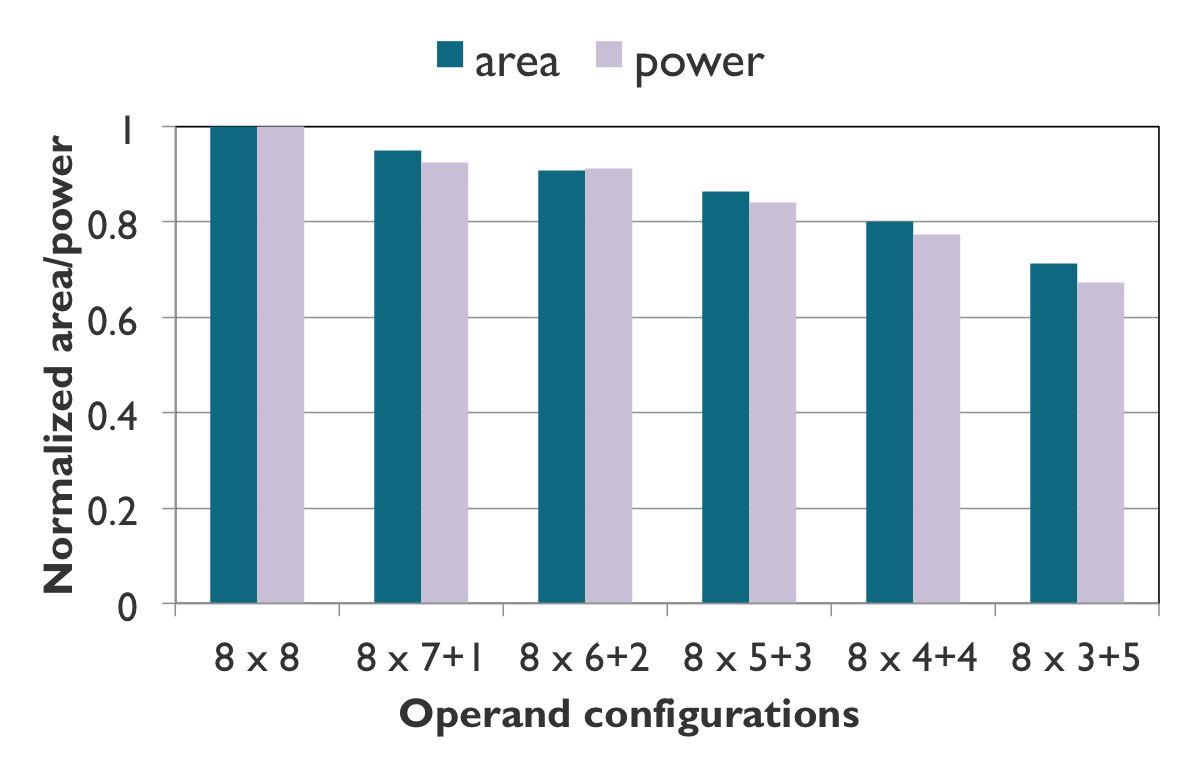}
\caption{
Hardware implementation trade-off results for different multiplier
configurations. The area and power are normalized
with respect to the case with two 8-bit fixed-point operands.
The floating-point operands are represented as mantissa bit-width +
exponent bit-width.
}
\label{fig:hw_tradeoff}
\end{figure}

To validate the claim in Section~\ref{subsec:implementation_consideration}, 
we implement the multiplier with different operand 
configurations using a commercial 16nm process technology and libraries.
The results are highlighted in Fig.~\ref{fig:hw_tradeoff}.
The floating-point configuration is represented
as $m$+$e$, i.e., mantissa bit-width + exponent bit-width.
Here we assume the baseline is a multiplier with two 8-bit fixed-point operands,
i.e., 8$\times$8. 
The area and power numbers are all normalized with respect to the
baseline.
With the same bit-width, the proposed scheme of combining fixed-point and 
floating-operands can reduce both area and power.
The reduction increases with less mantissa and more exponent bits, as a
shifter is more efficient than a multiplier.
As an example, the floating-point operand with 4-bit mantissa and 4-bit 
exponent, i.e. 8x4+4, can reduce the power and area by more than 20\%,
compared to 8x8 fixed-point multiplication.

\section{Experimental Results}
\label{sec:results}

As discussed in Section~\ref{subsec:range_vs_precision}, we perform the weight 
quantization based on {\em{Caffe}}~\cite{jia2014caffe}. We denote the representation as
($m$, $e$), where $m$ is the mantissa bit-width and $e$ is the exponent
bit-width. $e$=0 means fixed-point representation.

\begin{table}
\small
\caption{AlexNet accuracy with different representations.
$e$=0 means fixed-point representation. The floating-point
number is with the implicit bit.
}
\label{tbl:alex_accuracy}
\begin{tabular}{|c|c|c|c|c|c|c|}
\hline
  & $e$=0 & $e$=1 & $e$=2 & $e$=3 & $e$=4 & $e$=5 \\
\hline
$m$=1 & 0.002 & 0.002 & 0.003 & 0.920 & 0.918 & 0.918 \\
\hline
$m$=2 & 0.003 & 0.003 & 0.003 & 0.983 & 0.982 & 0.982 \\
\hline
$m$=3 & 0.002 & 0.003 & 0.003 & 0.995 & 0.995 & 0.995 \\
\hline
$m$=4 & 0.016 & 0.003 & 0.003 & 0.997 & 1.001 & 1.001 \\
\hline
$m$=5 & 0.775 & 0.002 & 0.003 & 0.994 & 0.999 & 0.998 \\
\hline
$m$=6 & 0.979 & 0.002 & 0.003 & 0.995 & 0.999 & 0.999 \\
\hline
$m$=7 & 0.996 & 0.002 & 0.003 & 0.995 & 0.999 & 0.999 \\
\hline
$m$=8 & 0.998 & 0.002 & 0.003 & 0.995 & 1.000 & 1.000 \\
\hline 
$m$=9 & 1.001 & 0.002 & 0.003 & 0.995 & 1.000 & 1.000 \\
\hline
$m$=10& 0.999 & 0.002 & 0.003 & 0.995 & 1.001 & 1.001 \\
\hline
\end{tabular}
\end{table}


We evaluate the network accuracy with different bit-width setting.
Table~\ref{tbl:alex_accuracy} highlights some results for AlexNet.
The network is non-functioning (i.e., with close to 0 accuracy)
when $e$ is small, i.e., with limited representation range.
To achieve full accuracy, AlexNet requires 8 bits for fixed-point
representation, i.e., (7, 0) with 1 sign bit, or 7 bits for floating-point 
representation with
(3, 3) configuration.
If the implementation only targets AlexNet, the proposed
number representation can achieve 12.5\% weight storage reduction and
8\% power reduction in multiplication.
The benefit will increase for CNNs that require more fixed-point
bit-width.

As discussed earlier, one of the requirement for number representation
scheme is the consistency across different networks.
This is especially important for the hardware implementation
to be future-proof and viable for different CNN models.
Some results of the normalized accuracy of different network are
highlighted in Table~\ref{tbl:network_accuracy}.
The 7-bit fixed-point configuration used for AlexNet also works
for SqueezeNet, but is not adequate for GoogLeNet and VGG-16.
10-bit fixed-point representation is required to get consistent 
accuracy across all networks used in this study. 

\begin{table}
\small
\caption{Normalized Accuracy for different networks.}
\label{tbl:network_accuracy}
\begin{tabular}{|c|c|c|c|c|}
\hline
 ($m$, $e$) & AlexNet & SqueezeNet & GoogLeNet & VGG-16 \\
\hline
 (7, 0)     & 1.00    & 1.00       & 0.85   & 0.02 \\
\hline
 (10, 0)    & 1.00    & 0.99       & 0.99   & 1.00 \\
\hline
 (3, 4)     & 0.99    & 0.99       & 0.99   & 1.00 \\
\hline
\end{tabular}
\end{table}


By using proposed number representation scheme, we only need 7-bit
floating-point, i.e, (3, 4) configuration. Therefore, we can replace 
11-bit weights (10-bit fixed-point number plus sign bit) with 8-bit weights
(3-bit mantissa, 4-bit exponent and 1 sign bit).
This results in 36\% storage reduction for weights and 50\% power reduction
in multiplication.

\section{Conclusion}
\label{sec:conclusion}

In this work, we propose CNN inference implementation with
floating-point weights and fixed-point activations. We give
the motivation for the proposed number representation scheme from
both algorithmic and hardware implementation perspectives. 
The proposed scheme can reduce the weight storage by up to 36\%
and the multiplier power by up to 50\%. 
Future work will investigate the impacts of network topology
and training on the number representation requirements.

\bibliographystyle{icml2015}

\end{document}